\documentclass{article}
\usepackage{spconf,amsmath,graphicx}
\usepackage{makecell}
\usepackage{epsfig}
\usepackage{amssymb}
\usepackage{pdflscape}
\usepackage{tabularx}
\usepackage{lipsum}
\usepackage{multicol}
\usepackage{multirow}
\usepackage{array}
\usepackage{stfloats}
\usepackage{algorithm}
\usepackage{algpseudocode}

\title{A New Perspective for Flexible Feature Gathering in Scene Text Recognition Via Character Anchor Pooling}

\name{Shangbang Long\textsuperscript{$*\dagger$}\thanks{* These authors have contributed equally to this work.}, Yushuo Guan\textsuperscript{$*\dagger$}, Kaigui Bian\textsuperscript{$\dagger$} and Cong Yao\textsuperscript{\P}}

\address{\textsuperscript{$\dagger$}Peking University, Beijing, China\\
\textsuperscript{\P}Megvii (Face++) Technology Inc.}

\begin{document}
%
\maketitle

\begin{abstract}
   Irregular scene text recognition has attracted much attention from the research community, mainly due to the complexity of shapes of text in natural scene. 
   However, recent methods either rely on shape-sensitive modules such as bounding box regression, or discard sequence learning. 
   To tackle these issues, we propose a pair of coupling modules, termed as \textbf{Character Anchoring Module (CAM)} and \textbf{Anchor Pooling Module (APM)}, to extract high-level semantics from two-dimensional space to form feature sequences. 
   The proposed CAM localizes the text in a shape-insensitive way by design by anchoring characters individually. 
   APM then interpolates and gathers features flexibly along the character anchors which enables sequence learning. 
   The complementary modules realize a harmonic unification of spatial information and sequence learning. 
   With the proposed modules, our recognition system surpasses previous state-of-the-art scores on irregular and perspective text datasets, including, ICDAR 2015, CUTE, and Total-Text, while paralleling state-of-the-art performance on regular text datasets.
\end{abstract}

\section{Introduction}
Scene text recognition has been an increasingly popular research topic in computer vision in the last few decades. 
As a carrier for information of high-level semantics, the ability to read text from natural images is beneficial for the understanding of the surrounding scenes through computer vision~\cite{long2018scene}. 
There are various applications, including \textit{instant translation}, 
\textit{robot navigation}, 
\textit{industrial automation}, 
and \textit{traffic sign reading for autonomous vehicles}. 
More recently, the detection and recognition of irregular text, e.g., text arranged in a curved line, has attracted much attention~\cite{shi2016robust,long2018textsnake}. 

As deep learning is widely applied to this field, most recent methods follow an encoder-decoder framework, in which, images are discomposed as a sequence of pixel frames, starting from the left side of the image to the right~\cite{shi2016robust,cheng2017arbitrarily,yin2017scene,yang2017learning}. 
The framework can be summarized and termed as \textit{Convolutional Recurrent Neural Networks (CRNN)}, where convolutional neural network (CNN) layers  encode the image into deep features and compress the height of feature maps into $1$.
The compressed feature has a size of $C \times 1 \times W$, and therefore is processed as a $C$-dimensional sequential features with $w$ time steps, which are then further encoded and decoded by a Seq2Seq model~\cite{sutskever2014sequence}. 
We refer to this feature encoding process as \textbf{feature gathering}, which transforms 2D images into 1D feature sequences. 
These methods produce good results when the text in the image is horizontal. 

For oriented and irregular text, the use of a rectification layer~\cite{shi2016robust,shi2018aster} based on Spatial Transformer Networks~\cite{jaderberg2015spatial} alleviates the problem to some extent. 
The rectification layer first predicts a bounding polygon to precisely locate the text, then generates grids according to the polygon, and finally transforms it. 
The idea behind is intuitive and has proven effective. However, the coordinates of bounding polygons are predicted via fully connected networks, and fail when the text has a shape that is poorly represented in the training dataset.  
The fact that the polygon prediction is shape-sensitive and may not generalize well to unseen shapes limits the potential of rectification-based methods. 
Similar problem also exists in 2D attention method~\cite{li2018show}, which is proven by a less competent score on blurred datasets. 

Recently, CA-FCN~\cite{liao2019two} takes the two-dimensional spatial distribution of text into consideration, and text recognition is reformulated as semantic segmentation, where character categories are segmented from the background. 
However, their method abandons the use of recurrent neural networks (RNN), and thus fails to obtain an overall vision. 
Therefore, it is prone to missing characters and performance drops significantly especially when the text are blurred. 

To tackle the challenges mentioned above, we design a flexible feature gathering method which deeply integrates the idea of sequence learning and the idea of considering two-dimensional spatial distribution. 

The key step is to gather the feature vectors from the shared feature maps along \textit{character anchor line (CAL)}, to form sequential features for subsequent sequence-to-sequence learning. 
To achieve this function, we design two novel modules, \textbf{Character Anchoring Module (CAM)} that anchors characters and \textbf{Anchor Pooling Module (APM)} that forms sequential features. 
The CAM module detects character center anchors individually in the form of heat map, which is inherently shape-agnostic, and therefore adapts better to irregular text even with unseen shapes. 
Then, the APM module interpolates the feature vectors along the CAL, and gathers into a sequence. 
Instead of only considering detected character centers, the interpolation can achieve robustness against challenging image conditions by filling missing characters. 
While the shape-insensitive CAM can robustly guide the APM along the text features to form sequence, APM can correct errors made by CAM. 
Based on these two complementary modules, we propose a recognition model, termed as \textbf{Character Anchor Pooling Network (CAPNet)}.
Compared with previous methods, our method successfully and harmonically unifies shape-agnostic localization and sequence learning. 
 Our methodology is demonstrated in Fig. \ref{fig:Pipeline}.


The contributions of our paper can be summarized as follows. (1) We propose two innovative and complementary modules, \textbf{Character Anchoring Module} and \textbf{Anchor Pooling Module}, to successfully harmonize sequence learning and two-dimensional spatial arrangement of text. 
(2) Empirically, the proposed \textbf{CAPNet} outperforms previous state-of-the-art results on irregular and perspective datasets, including ICDAR 2015, CUTE and Total-Text.
It also outperforms previous methods on several horizontal text dataset such as IIIT5K and ICDAR 2013, while paralleling on other datasets.
(3) We provide in-depth qualitative and quantitative analysis as well as ablation tests to further understand its strengths and dependencies. 

\section{Methodology}\label{sec-model}

\subsection{Pipeline}
\begin{figure}
\centering
\includegraphics[width=0.8\columnwidth]{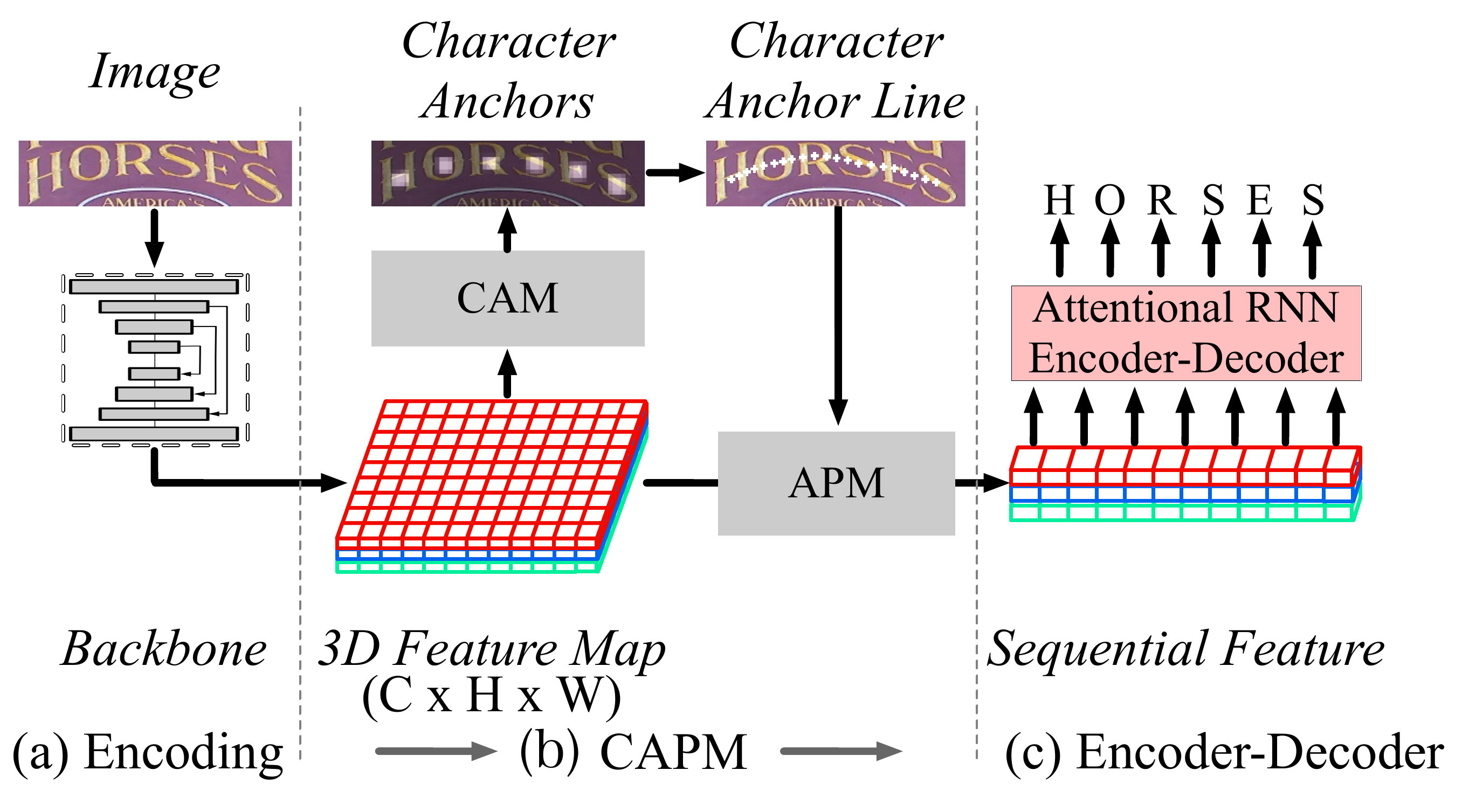}
\vspace{-3mm}
\caption{The pipeline of the proposed CAPNet: 
\textit{(a)}: ResNet50+FPN backbone; 
\textit{(b)}: the feature gathering module that we refer to as Character Anchor Pooling Module; 
\textit{(c)}: the sequence learning module. We apply an attentional RNN-based encoder-decoder network.}  
\label{fig:Pipeline}
\vspace{-4mm}
\end{figure}

As shown in Fig. \ref{fig:Pipeline}, the pipeline of CAPNet contains the following components: (1) A fully convolutional backbone module for feature encoding, producing shared feature maps for the following steps. 
(2) The \textbf{Character Anchor Pooling Module (CAPM)} that comprises of CAM and APM.  
(3) An RNN-based recognition module that encodes the pooled features and decodes them into text symbols. 

\subsection{Model Architecture}
\noindent\textbf{Backbone:} We use a standard 50-layered ResNet~\cite{He_2017_Res} as backbone 
and FPN~\cite{Lin_2017_CVPRPyramid} connections to integrate features from different stages. 
The upsampling in the FPN branch is performed by bi-linear interpolation. 
Other settings follow ~\cite{Lin_2017_CVPRPyramid}. 
All CNN layers in FPN have $256$ filters. 
The kernel size is $1$ for lateral connections, and $3$ for top-down connections.
The network produces shared feature maps with $256$ channels, and is one-quarter of the size of the input image. 



\noindent\textbf{Character Anchoring Module:} To flexibly and robustly localize a text instance, we propose to anchor each individual character instead. 
Note that, in bounding polygon regression~\cite{shi2018aster} that localizes the text as an entirety, each control point will depend on the overall shape. 
This results in shape-sensitivity. 
On the contrary, the localization of characters does not have dependencies over the rest of the images. 
Therefore, detecting characters is insensitive to text shape by design. 
For easy separation of each character anchor, we define character anchors as shrank character boxes. 
The downsampling ratio is $\frac{1}{4}$.
The CAM contains a two-layered CNN, which we found strong enough to detect character centers. 
The first CNN layer has a kernel size of $3$ and $256$ filters in total. The second one is in essence a pixel-wise classification layer, which takes the shared feature maps as input, and produces a heat map, indicating the probability for each pixel to be the character anchor. 
The prediction map has the same size as the shared feature maps.



\begin{figure}
\centering
\includegraphics[width=0.8\linewidth]{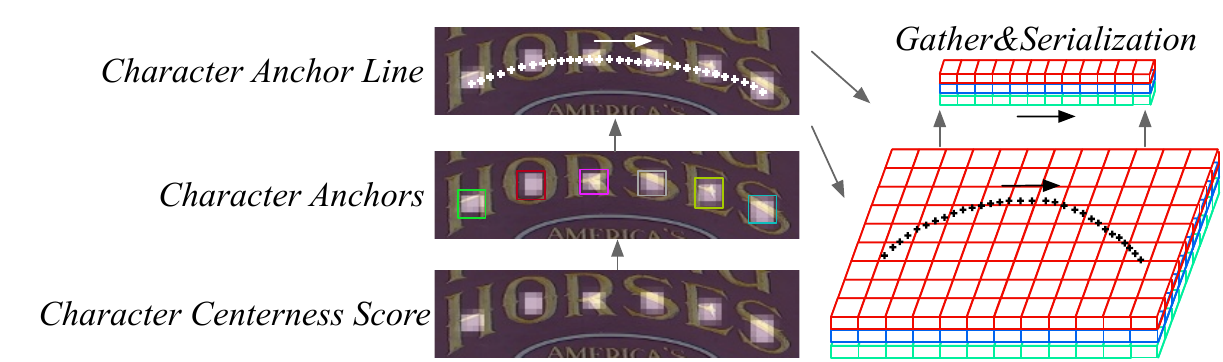}
\vspace{-3mm}
\caption{Anchor Pooling Module: 
(1) separating and grouping character center pixels to form individual character anchor; 
(2) sampling the CAL along generated character anchors;
(3) feature pooling along CAL, and concatenation into sequence.}  
\label{fig:pool}
\vspace{-4mm}
\end{figure}

\noindent\textbf{Anchor Pooling Module for Feature Gathering:} The details of APM is illustrated in Fig. \ref{fig:pool}. 
First, we separate and aggregate adjacent positive responses on the predicted character anchor heat map into groups, and each group would indicate one character anchor. 
Then we take the midpoint of each group and produce the coordinates of each character anchor. 
The coordinates are sorted from left to right, and the ordered list of sorted coordinates forms the basis of CAL. 
To enrich our feature sequence, we evenly sample a fixed number of markers along the sorted coordinates, which makes the CAL. 
The markers have floating point coordinates. 
We denote the number of markers in CAL as $M$. 
For each marker in CAL, we bi-linearly interpolate a $C\times 1\times 1$ feature vector from the corresponding floating point position on the shared feature maps. 
The last step is to concatenate the extracted features in order, and we obtain sequential features, with a size of $M\times C$, which has $M$ time steps. 

Besides, experiments results show that extracting features directly from the shared feature maps produced by backbone would result in worse performance. 
Therefore we add two extra CNN layers to further encode the shared feature maps. 
The character anchor pooling is performed on the further encoded feature maps, while the character anchoring still performs on the shared feature maps. 
Most previous works follow similar practice~\cite{shi2018aster}. 
This is mainly because localization and recognition require different aspects of the features. 
Besides, the two additional CNN layers also enlarge the receptive field to cover more visual features. 


\noindent\textbf{Attentional Encoder-Decoder:} The recognition module is an RNN-based attentional encoder-decoder. 
The encoder is a one-layered bidirectional LSTM. It encodes the features extracted by character anchor pooling and outputs $H=[h_1,...,h_M]$. The encoder captures long-term dependencies and maintains an overall vision over the pooled sequential features. This is important especially when individual characters are blurred, or missing. 
The size of the hidden state is set as $256$ for each direction. 
The hidden states of both directions are concatenated. 

For the decoder, we use a unidirectional LSTM. 
The size of the hidden state is $512$. 
The hidden-state at time step $0$ is initialized to equal the last hidden state of the encoder. 
The decoder is equipped with the attention mechanism~\cite{bahdanau2014neural}. It consists of an alignment module and a RNN-decoder module. At time step $i$, the alignment module calculates the attention weights $\alpha_i$, which effectively indicate the importance of every item of the encoder output $H$: 
\begin{equation}
    \alpha_{i j}=\frac{\exp \left(e_{i j}\right)}{\sum_{m=1}^{M} \exp \left(e_{i m}\right)}, e_{i j}=a\left(s_{i-1}, h_{j}\right)
\end{equation}

\noindent The RNN module produces an output, $x_i$, and a new state $s_i$:
\begin{equation}
    (x_i, s_i) = \text{LSTM}(s_{i-1}, \hat{y}_{i-1}, \sum_{j=1}^{T_{x}} \alpha_{i j} h_{j})
\end{equation}

\noindent Given the alphabet set $\Psi$, the decoder takes $x_i$ as input and predicts the output symbol $\hat{y}_i$ with softmax:
\begin{align}
    [p_i^1, ..., p_i^{|\Psi|}] &= \text{softmax}\left(W_{o} x_i +b_o\right) \\
    \hat{y}_i &= \text{argmax}_{\psi} p_t^{\psi}
\end{align}

\subsection{Training Targets} 
The network is trained jointly to match the ground-truth labels of character centers and text sequence. Character anchor heat map is a binary $0/1$ prediction at the pixel level. Given the number of pixels in the feature maps as $N$, the loss is defined as the following binary cross entropy loss:

\begin{equation}
\small
\centering
\begin{aligned}
L_{loc} &= -\frac{1}{N}\sum_{n=1}^N\{y_nlog[p_{\theta}({y_n}|Image)]\\
 &+ (1-y_n)log[1-p_{\theta}({y_n}|Image)]\}
\end{aligned}
\label{eq-loss-det}
\end{equation}

\noindent The recognition module predicts a symbol sequence. 
We denote the ground-truth of the text sequence as $\{y_{1}, ..., y_{T}\}$. 
In training, we pad the sequence with \textit{end-of-sentence (\textit{EOS})} symbol due to variable lengths. 
The loss is defined as negative maximum-likelihood averaged over time $T$:

\begin{equation}
\small
\centering
L_{rec} = -\frac{1}{T}\sum_{i=1}^T log[p(y_{i}|Image)]
\label{eq-loss-rec}
\end{equation}

\noindent The training target is the weighted sum of the localization loss and recognition loss:

\begin{equation}
\small
\centering
L = \lambda_1 L_{loc} + \lambda_2 L_{rec}
\label{eq-loss}
\end{equation}

\noindent where $\lambda_1$ and $\lambda_2$ are weights and are set to $1.0$ by default in our experiments.

\section{Experiment}

\textbf{Datasets:} Following previous works, we train our network solely on synthetic datasets, SynthText~\cite{gupta2016synthetic} and Synth90K~\cite{jaderberg2014synthetic}. There are $7M$ image crops in SynthText, which are annotated at the character level. Synth90K contains 9M greyscale images. It has a balanced distribution
over a 90K-vocabulary, and only has annotations of
ground-truth word.

We evaluate the trained network on various real-world datasets:
IIIT5K~\cite{mishra2012scene}, SVT~\cite{wang2010word}, SVT-Perspective~\cite{quy2013recognizing}, IC03~\cite{lucas2003icdar}, IC13~\cite{karatzas2013icdar}, IC15, CUTE~\cite{risnumawan2014robust} and Total-Text~\cite{kheng2017total}.



\begin{table*}[h]
\caption{Performance of different methods over $8$ datasets. ``50'', ``1K'', ``Full'' are the size of lexicons. ``0'' means no lexicon.  ``90K'' and ``ST'' are the Synth90k and the SynthText datasets, respectively. ``ST$^{+}$'' means including character-level annotations. ``Private'' means private training data.
``-'' means the score is not reported in the paper. 
}
\label{sota-recognition}
\renewcommand{\arraystretch}{1.4}
\centering
\resizebox{\textwidth}{!}{%
\begin{tabular}{|c|c|ccc|cc|ccc|c|c|c|c|c|}
\hline 
\multirow{2}{*}{Methods} & \multirow{2}{*}{Backbone, Data} & \multicolumn{3}{c|}{IIIT5K} & \multicolumn{2}{c|}{SVT} & \multicolumn{3}{c|}{IC03} & IC13 & IC15 & SVT-P & CUTE & Total-Text \tabularnewline
\cline{3-15} 
& &  50 & 1k & 0 & 50 & 0 & 50 & Full & 0 & 0 & 0 & 0 & 0 & 0\tabularnewline
\hline 
Aster~\cite{shi2018aster} & ResNet, 90K+ST & {99.6} & {98.8} & {93.4} & {97.4} & {{89.5}} & 98.8 & \textbf{98.0} & {94.5} & 91.8 & {76.1} & {78.5} & {79.5} & -\tabularnewline
CA-FCN~\cite{liao2019two} & Attentional VGG, ST$^{+}$ + extra ST  & {99.8} & {\textbf{98.9}} & {92.0} & {98.8} & {86.4} & - & - & {-} & 91.5 & {-} & {-} & {79.9} & 61.6 \tabularnewline
2D Attn~\cite{li2018show} & ResNet, ST + 90K + extra ST & - & - & 91.5 & - & 84.5 & - & - & - & 91.0 & 69.2 & 76.4 & 83.3 & -\tabularnewline
\hline
\textbf{CAPNet} & ResNet, 90K+ST$^{+}$ & {\textbf{99.8}} & {98.8} & {\textbf{93.7}} & {\textbf{98.9}} & {88.9} & \textbf{99.3} & 97.8 & {\textbf{94.6}} & \textbf{92.4} & {\textbf{76.6}} & 78.8 & {\textbf{86.8}} & \textbf{62.7}\tabularnewline
\hline
\end{tabular}
}
\label{tbl:comparison-to-sota}
\vspace{-7mm}
\end{table*}

\begin{table}
\caption{Results of ablation tests compared with CAPNet. }
\label{bl-recognition}
\centering
\scalebox{0.67}{
\begin{tabular}{|c|c|c|c|c|c|c|c|c|}
\hline 
Methods & IIIT5K & SVT & IC03 & IC13 & IC15 & SVT-P & CUTE & Total \tabularnewline
\hline 
\makecell{CA-FCN + RNN} & {92.1} & {87.0}& {93.9} & 91.6 & {71.3} & {72.2} & {{80.6}} & 62.0\tabularnewline
\hline
CAPNet + VGG & {93.1} & {{87.5}}& {{94.3}} & {92.1} & {74.9} & {{77.1}} & {{86.3}} & {\textbf{62.9}}\tabularnewline
\hline 
\textbf{CAPNet} & {{\textbf{93.7}}} & {\textbf{88.9}} & {\textbf{94.6}} & \textbf{92.4} & {\textbf{76.6}} & {\textbf{78.8}} & {{\textbf{86.8}}} & 62.7\tabularnewline
\hline 
\end{tabular}
}
\vspace{-5mm}
\end{table}

\noindent\textbf{Implementation:} 
We use the two-staged strategy to jointly train the localization and recognition module.
In the first training stage, character anchor pooling is performed using ground-truth character centers. 
This stage requires character-level annotations, and therefore the network is only trained on SynthText, which provides effortless character-level annotations. In the second stage, character anchor pooling is performed using the output produced by the CAPM. In this stage, the network is trained on both SynthText and Synth90K and learns to calibrate its own outputs. The localization loss is only computed and averaged over SynthText data, and the recognition loss is computed with both. We step into the second stage of training once the moving average of $L_{loc}$ falls below $0.35$ for the first time. We train $4$ epochs in total. The learning rate is set to $1.0$ for the first epoch, and decays to $10^{-1}$ at the second epoch. At the last two epochs, the learning rate is set to $10^{-2}$.

The network is implemented with PyTorch. The length of a pooled feature sequence is set to $36$ to match the length of feature sequences in most previous works. All words are padded to $64$ tokens with the special symbol \textit{EOS} in training. The recognition module recognizes ten digits and 26 case-insensitive alphabets. During training and evaluation, images are resized to $64\times 256$.  For data augmentation, we apply random Gaussian noise and motion blur. To train our network, we use the ADADELTA~\cite{zeiler2012adadelta} optimizer with default parameters to mini-batches of randomly selected $512$ samples. All experiments are performed on $4$ NVIDIA GeForce 1080 Ti GPU, each with 12GB memory.

\noindent \textbf{Performance on Straight Text:} We achieve better performance on $4$ of the $6$ straight text datasets, including IIIT5K ($+0.3$), IC03 ($+0.1$), and IC15 ($+0.5$). 
We also parallel previous methods on other straight text datasets, including SVT ($88.9/98.9$) and IC13 ($92.4$). Results are shown in Tab. \ref{sota-recognition}. Therefore, our method is no worse than previous methods in recognizing straight text and even better on some datasets. 

\noindent \textbf{Performance on Curved Text:} As for curved dataset, we outperforms previous state-of-the-art method using rectification~\cite{shi2018aster} by an absolute improvement of $5\%$ on CUTE. CAPNet also achieves higher score than the 2D attention baseline~\cite{li2018show} by $3.5\%$ on CUTE, while surpassing by $7.2\%$ on IC15 and $2.4\%$ on SVT-P. 
The superior performance verifies the effectiveness of our method. 
For more comprehensive comparison, we also evaluate our method on Total-Text, a large curved text dataset containing $6\times$ more data than CUTE. 
Our method still outperforms previous SOTA result by $1.1\%$. 

\noindent\textbf{Ablation Study} We makes two variants of CAPNet by: (1) Change the backbone of CAPNet from ResNet-50 to VGG-16. (2) Add the same RNN module of CAPNet into CA-FCN. As shown in Table~\ref{bl-recognition}, CAPNet outperforms the two variants, which shows that ResNet-50 is more suitable than VGG-16 in CAPNet and the improvements not only comes from the sequence learning but also the combination of CAM and APM. 

\noindent\textbf{The Quality of the Predicted CAL:} To estimate the quality of the predicted CAL quantitatively, we consider the correlation coefficients of $x$ and $y$ coordinates between predicted CAL and ground-truth CAL. The higher the correlation, the better control points CAPM produces.

\begin{figure}[t]
\centering
\includegraphics[width=0.85\linewidth]{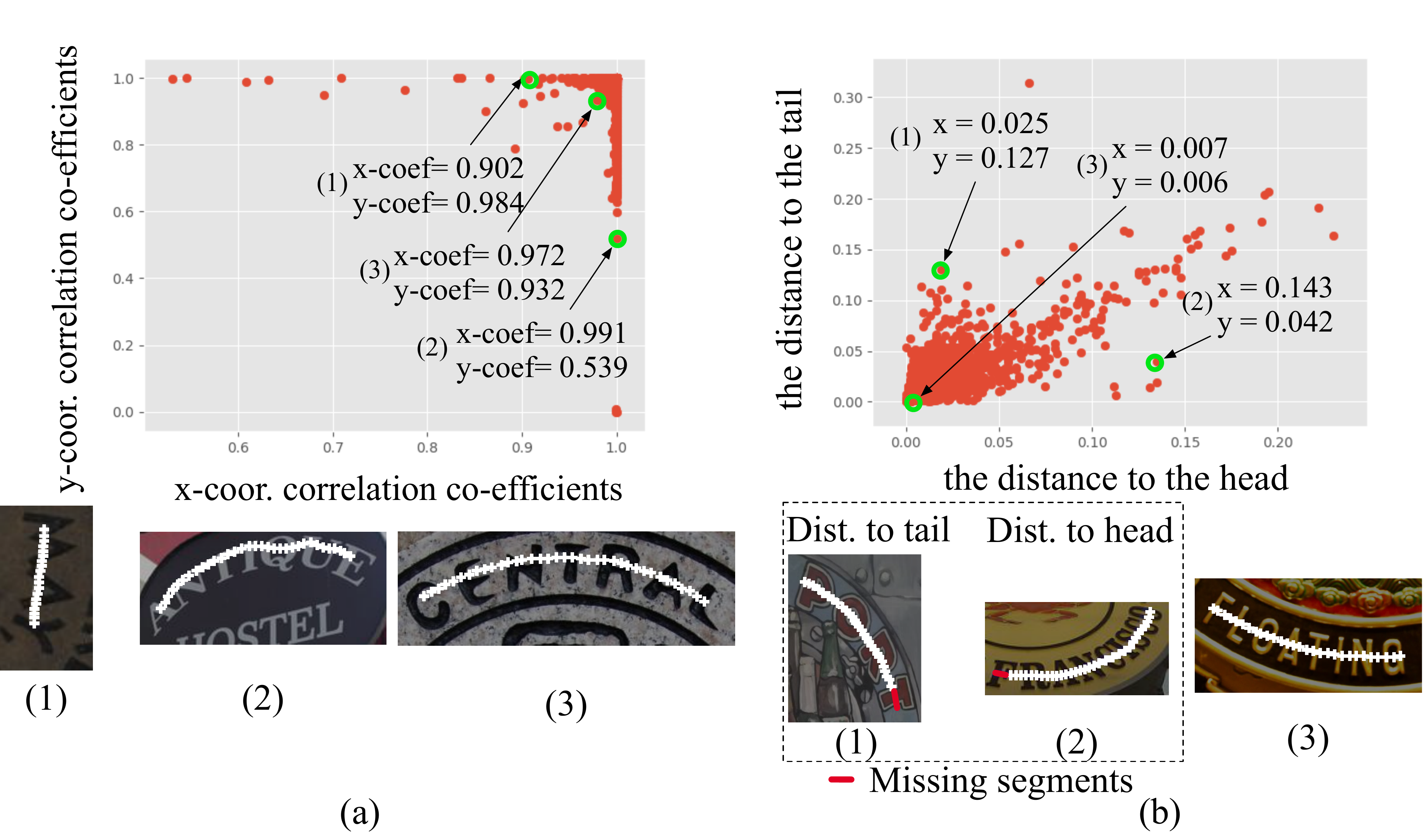}
\vspace{-4mm}
\caption{(a): Correlation coefficients of $x\&y$ over all images in Total-Text, and samples marked in the diagram. (b): Distance (length of missing segments) from the CAL to the text head and tail (marked in red).}  
\label{fig:acc}
\vspace{-4mm}
\end{figure}

We evaluate on Total-Text, which provides word bounding polygons to compute ground-truth CAL.
We compute CAL correlation coefficients for each image, and draw a 2D scatter diagram of all test samples. The diagram is shown in Fig. \ref{fig:acc}. In (a), we give an illustrative description of how character anchors look like given different levels of correlation.
Nearly all scatter points fall on the top-right corner of the diagram which indicates that, for most images, CAPM performs good enough to accurately sketch the shape of the text. 
In (b), we analyse the distances from the predicted CAL to the text head and tail regions. Most scatter points fall on the bottom-left corner of the diagram. The observations  verify that predicted CALs are flawless in most cases.

\vspace{-1ex}
\section{Conclusion}
\vspace{-1ex}
In this paper, we investigate how to realize sequence learning in two-dimensional space for text recognition, and propose Character Anchor Pooling Module (CAPM). CAPM pools features along character anchors, which is formed based on the localization of character centers. 
Our proposed CAPM localizes text flexibly, and provides a better basis for the subsequent sequence learning. 
The localization module is shape-agnostic, and therefore can produce accurate outputs on curved text even if it is only trained on mostly straight text.
In experiments, we find that our method outperforms existing methods, on both straight and curved text datasets, which demonstrates the effectiveness of our method.
We also perform in-depth analysis with regard to our CAM, to show that it is good enough even under some difficult situations.
In conclusion, our paper makes an effort attempting to find proper representation for the irregular scene text recognition. 

\clearpage

\bibliographystyle{IEEEbib}
\bibliography{refs}

\begin{thebibliography}{10}

\bibitem{long2018scene}
Shangbang Long, Xin He, and Cong Yao,
\newblock ``Scene text detection and recognition: The deep learning era,''
\newblock {\em arXiv preprint arXiv:1811.04256}, 2018.

\bibitem{shi2016robust}
Baoguang Shi, Xinggang Wang, Pengyuan Lyu, Cong Yao, and Xiang Bai,
\newblock ``Robust scene text recognition with automatic rectification,''
\newblock in {\em CVPR 2016}, 2016.

\bibitem{long2018textsnake}
Shangbang Long, Jiaqiang Ruan, Wenjie Zhang, Xin He, Wenhao Wu, and Cong Yao,
\newblock ``Textsnake: A flexible representation for detecting text of
  arbitrary shapes,''
\newblock in {\em ECCV}, 2018.

\bibitem{cheng2017arbitrarily}
Zhanzhan Cheng, Xuyang Liu, Fan Bai, Yi~Niu, Shiliang Pu, and Shuigeng Zhou,
\newblock ``Arbitrarily-oriented text recognition,''
\newblock {\em CVPR 2018}, 2017.

\bibitem{yin2017scene}
Fei Yin, YiChao Wu, XuYao Zhang, and ChengLin Liu,
\newblock ``Scene text recognition with sliding convolutional character
  models,''
\newblock {\em arXiv preprint arXiv:1709.01727}, 2017.

\bibitem{yang2017learning}
Xiao Yang, Dafang He, Zihan Zhou, Daniel Kifer, and C~Lee Giles,
\newblock ``Learning to read irregular text with attention mechanisms,''
\newblock in {\em IJCAI}, 2017.

\bibitem{sutskever2014sequence}
Ilya Sutskever, Oriol Vinyals, and Quoc~V Le,
\newblock ``Sequence to sequence learning with neural networks,''
\newblock in {\em NIPS}, 2014.

\bibitem{shi2018aster}
Baoguang Shi, Mingkun Yang, XingGang Wang, Pengyuan Lyu, Xiang Bai, and Cong
  Yao,
\newblock ``Aster: An attentional scene text recognizer with flexible
  rectification,''
\newblock {\em IEEE transactions on pattern analysis and machine intelligence},
  vol. 31, no. 11, pp. 855--868, 2018.

\bibitem{jaderberg2015spatial}
Max Jaderberg, Karen Simonyan, Andrew Zisserman, et~al.,
\newblock ``Spatial transformer networks,''
\newblock in {\em Advances in neural information processing systems}, 2015.

\bibitem{li2018show}
Hui Li, Peng Wang, Chunhua Shen, and Guyu Zhang,
\newblock ``Show, attend and read: A simple and strong baseline for irregular
  text recognition,''
\newblock {\em AAAI}, 2019.

\bibitem{liao2019two}
Minghui Liao, Jian Zhang, Zhaoyi Wan, Fengming Xie, Jiajun Liang, Pengyuan Lyu,
  Cong Yao, and Xiang Bai,
\newblock ``Scene text recognition from two-dimensional perspective.,''
\newblock in {\em AAAI}, 2019.

\bibitem{He_2017_Res}
Kaiming He, Xiangyu Zhang, Shaoqing Ren, and Jian Sun,
\newblock ``Deep residual learning for image recognition,''
\newblock in {\em CVPR}, 2016.

\bibitem{Lin_2017_CVPRPyramid}
Tsung-Yi Lin, Piotr Dollar, Ross Girshick, Kaiming He, Bharath Hariharan, and
  Serge Belongie,
\newblock ``Feature pyramid networks for object detection,''
\newblock in {\em CVPR}, 2017.

\bibitem{bahdanau2014neural}
Dzmitry Bahdanau, Kyunghyun Cho, and Yoshua Bengio,
\newblock ``Neural machine translation by jointly learning to align and
  translate,''
\newblock {\em ICLR 2015}, 2015.

\bibitem{gupta2016synthetic}
Ankush Gupta, Andrea Vedaldi, and Andrew Zisserman,
\newblock ``Synthetic data for text localisation in natural images,''
\newblock in {\em CVPR}, 2016.

\bibitem{jaderberg2014synthetic}
Max Jaderberg, Karen Simonyan, Andrea Vedaldi, and Andrew Zisserman,
\newblock ``Synthetic data and artificial neural networks for natural scene
  text recognition,''
\newblock {\em arXiv preprint arXiv:1406.2227}, 2014.

\bibitem{mishra2012scene}
Anand Mishra, Karteek Alahari, and CV~Jawahar,
\newblock ``Scene text recognition using higher order language priors,''
\newblock in {\em BMVC}, 2012.

\bibitem{wang2010word}
Kai Wang and Serge Belongie,
\newblock ``Word spotting in the wild,''
\newblock in {\em In Proceedings of European Conference on Computer Vision
  (ECCV)}. Springer, 2010, pp. 591--604.

\bibitem{quy2013recognizing}
Trung Quy~Phan, Palaiahnakote Shivakumara, Shangxuan Tian, and Chew Lim~Tan,
\newblock ``Recognizing text with perspective distortion in natural scenes,''
\newblock in {\em Proceedings of the IEEE International Conference on Computer
  Vision (ICCV)}, 2013, pp. 569--576.

\bibitem{lucas2003icdar}
Simon~M Lucas, Alex Panaretos, Luis Sosa, Anthony Tang, Shirley Wong, and
  Robert Young,
\newblock ``Icdar 2003 robust reading competitions,''
\newblock IEEE, 2003.

\bibitem{karatzas2013icdar}
Dimosthenis Karatzas, Faisal Shafait, Seiichi Uchida, Masakazu Iwamura,
  Lluis~Gomez i~Bigorda, Sergi~Robles Mestre, Joan Mas, David~Fernandez Mota,
  Jon~Almazan Almazan, and Lluis~Pere de~las Heras,
\newblock ``Icdar 2013 robust reading competition,''
\newblock in {\em Document Analysis and Recognition (ICDAR), 2013 12th
  International Conference on}. IEEE, 2013, pp. 1484--1493.

\bibitem{risnumawan2014robust}
Anhar Risnumawan, Palaiahankote Shivakumara, Chee~Seng Chan, and Chew~Lim Tan,
\newblock ``A robust arbitrary text detection system for natural scene
  images,''
\newblock {\em Expert Systems with Applications}, vol. 41, no. 18, pp.
  8027--8048, 2014.

\bibitem{kheng2017total}
Chee~Kheng Ch'ng and Chee~Seng Chan,
\newblock ``Total-text: A comprehensive dataset for scene text detection and
  recognition,''
\newblock in {\em ICDAR 2017}, 2017.

\bibitem{zeiler2012adadelta}
Matthew~D Zeiler,
\newblock ``Adadelta: an adaptive learning rate method,''
\newblock {\em arXiv preprint arXiv:1212.5701}, 2012.

\end{thebibliography}

\end{document}